\documentclass[sigconf, nonacm]{acmart}

\copyrightyear{2025}
\acmYear{2025}
\setcopyright{cc}
\setcctype{by}
\acmConference[RichMediaGAI '25]{Proceedings of the 3rd International Workshop on Rich Media With Generative AI}{October 27--28, 2025}{Dublin, Ireland}
\acmBooktitle{Proceedings of the 3rd International Workshop on Rich Media With Generative AI (RichMediaGAI '25), October 27--28, 2025, Dublin, Ireland}
\acmDOI{10.1145/3746262.3761976}
\acmISBN{979-8-4007-2044-4/2025/10}

\begin{document}

\title{$\mathtt{M^3VIR}$: A Large-Scale Multi-Modality Multi-View Synthesized \\ Benchmark Dataset for Image Restoration and Content Creation}


\author{Yuanzhi Li}
\affiliation{
  \institution{Santa Clara University}
  \city{Santa Clara}
  \state{CA}
  \country{USA}
}
\email{yli16@scu.edu}

\author{Lebin Zhou}
\affiliation{
  \institution{Santa Clara University}
  \city{Santa Clara}
  \state{CA}
  \country{USA}
}
\email{lzhou@scu.edu}

\author{Nam Ling}
\affiliation{
  \institution{Santa Clara University}
  \city{Santa Clara}
  \state{CA}
  \country{USA}
}
\email{nling@scu.edu}

\author{Zhenghao Chen}
\affiliation{
  \institution{University of Newcastle}
  \city{Callaghan}
  \state{NSW}
  \country{Australia}
}
\email{zhenghao.chen@newcastle.edu.au}

\author{Wei Wang}
\affiliation{
  \institution{Futurewei Technologies, Inc.}
  \city{San Jose}
  \state{CA}
  \country{USA}
}
\email{rickweiwang@futurewei.com}

\author{Wei Jiang}
\affiliation{
  \institution{Futurewei Technologies, Inc.}
  \city{San Jose}
  \state{CA}
  \country{USA}
}
\email{wjiang@futurewei.com}

\renewcommand{\shortauthors}{Yuanzhi Li et al.}

\begin{abstract}
The gaming and entertainment industry is rapidly evolving, driven by immersive experiences and the integration of generative AI (GAI) technologies. Training such models effectively requires large-scale datasets that capture the diversity and context of gaming environments. However, existing datasets are often limited to specific domains or rely on artificial degradations, which do not accurately capture the unique characteristics of gaming content. Moreover, benchmarks for controllable video generation remain absent. 

To address these limitations, we introduce $\mathtt{M^3VIR}$, a large-scale, multi-modal, multi-view dataset specifically designed to overcome the shortcomings of current resources. Unlike existing datasets, $\mathtt{M^3VIR}$ provides diverse, high-fidelity gaming content rendered with Unreal Engine 5, offering authentic ground-truth LR-HR paired and multi-view frames across 80 scenes in 8 categories. It includes $\mathtt{M^3VIR\_MR}$ for super-resolution (SR), novel view synthesis (NVS), and combined NVS+SR tasks, and $\mathtt{M^3VIR\_{MS}}$, the first multi-style, object-level ground-truth set enabling research on controlled video generation. Additionally, we benchmark several state-of-the-art SR and NVS methods to establish performance baselines. While no existing approaches directly handle controlled video generation, $\mathtt{M^3VIR}$ provides a benchmark for advancing this area. By releasing the dataset, we aim to facilitate research in AI-powered restoration, compression, and controllable content generation for next-generation cloud gaming and entertainment.
\end{abstract}




\maketitle

\section{Introduction}
\label{sec:intro}

The gaming and entertainment industry is undergoing rapid transformation, driven by advances in graphics, immersive experiences, and Generative AI (GAI) technologies that enable massive and easily accessible content creation. Modern users demand not only stunning, high-quality visuals but also fast and efficient media delivery to support truly immersive and interactive experiences.

However, the traditional video compression pipelines have difficulties in meeting such low-latency, high-quality demands. Most current solutions depend on heavy server-side computation and network delivery, with client devices functioning primarily as passive displays. Under bandwidth constraints, preventing input delays and excessive data transmission requires compressing high-quality frames aggressively, leading to degraded user experiences even on powerful client devices.  Traditional codecs such as H.264/H.265/H.266 \cite{vvc_std,hevc_std} or recent neural video coding methods \cite{DVC2019,FVC} cannot overcome this inherent bottleneck. These limitations highlight the need to rethink codec design and develop new compression frameworks specifically tailored for responsive, intelligent gaming and entertainment environments.

In the context of designing new codec frameworks, GAI can also play a crucial role. When applied to super-resolution (SR), image rendering, and content synthesis, GAI enables novel approaches to reduce transmission loads while maintaining visual quality. By offloading part of the rendering and reconstruction tasks to client devices, server-side computation and bandwidth requirements can be significantly alleviated. For example, servers can transmit only low-resolution (LR) frames, while client-side models reconstruct high-resolution (HR) outputs locally. In multiview scenarios such as immersive VR gaming, only a small subset of views needs to be transmitted, with the remaining views synthesized on the client side. This strategy, exemplified by technologies like NVIDIA's Deep Learning Super Sampling (DLSS) \cite{NVDLSS1,NVDLSS2,NVDLSS3}, demonstrates how GAI-driven frameworks can optimize the balance between bandwidth usage and device-side computational power. 

\begin{figure}[tbp]
\centering
    \includegraphics[width=\linewidth]{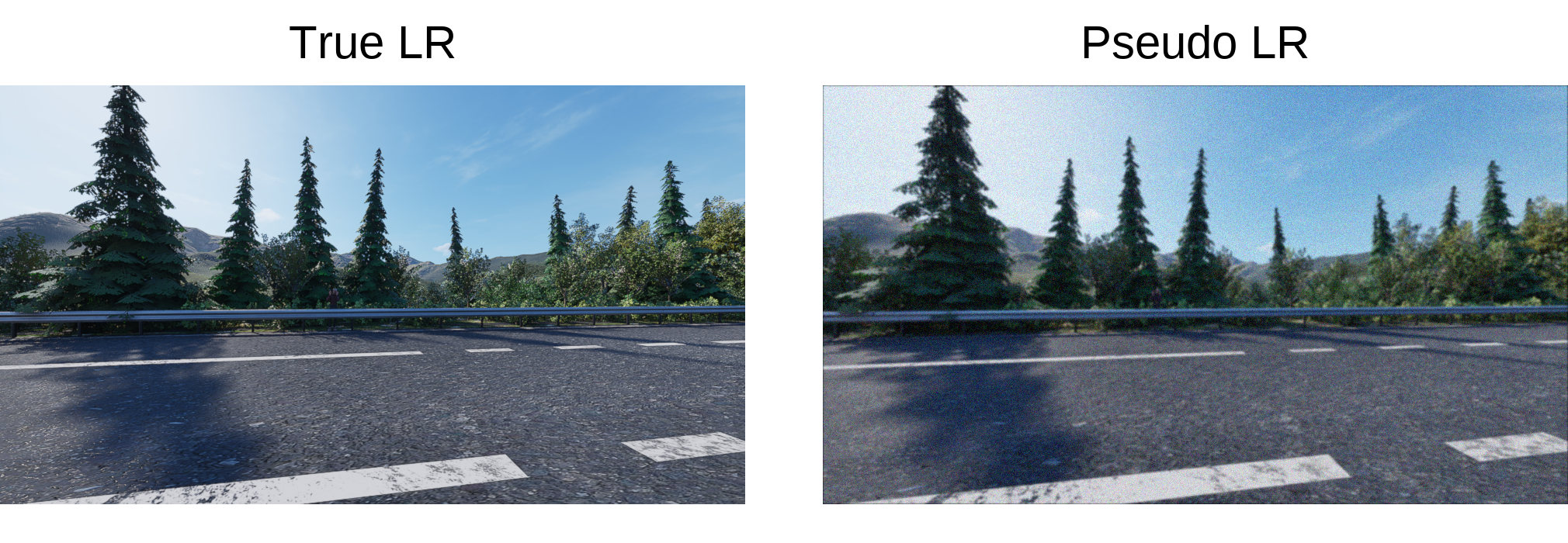}\vspace{-1em}
\caption{True LR vs. Pseudo LR (downsample+noise+blur)}\label{fig:lr_pseudo.png}\vspace{-1em}
\end{figure}

A key factor in the success of DLSS is the use of large-scale, high-quality ground-truth training data, such as the LR-HR paired frames or multiview gaming data, which closely align with true deployment scenarios. In contrast, much of the research community relies on pseudo training data for image restoration tasks \cite{IRreview1,IRreview2,IRreview3, VRreview1}. For instance, for SR tasks, LR images are typically generated by downsampling HR images and adding synthetic degradations like noise, blur, or compression artifacts.  Such artificially degraded data fails to represent real gaming content. As illustrated in Fig.~\ref{fig:lr_pseudo.png}, true LR gaming frames are often sharp and noise-free, lacking the degradations assumed in synthetic pipelines. Additionally, gaming visuals exhibit distinctive artifacts such as unnatural effects or object movements with minimal blur—unlike natural videos. These discrepancies highlight the necessity of using real ground-truth gaming data for effective model training.

Similarly, for generative content creation, benchmark datasets that reflect real-world characteristics are critical. While GAI shows great promise, its effectiveness is often limited by the lack of precise content control. Ensuring spatial-temporal consistency and physical accuracy remains challenging, especially when models rely solely on text prompts, which are inherently ambiguous for describing complex video content. Multi-modal guidance that combines visual references with textual descriptions offers a more reliable approach. Furthermore, unlike traditional pipelines where artists retain fine-grained control over assets, behaviors, and styles, generative models often produce content that is difficult to constrain. Therefore, high-quality, well-annotated benchmark datasets are essential to advancing controllable content generation.

In this work, we propose $\mathtt{M^3VIR}$, a large-scale, multi-modality, multi-view dataset with computer-synthesized virtual content, created to bridge the gap between existing generic datasets and the unique needs of gaming and entertainment restoration and content creation. 
Our objective is to enable the development of more effective visual restoration and generation methods specifically tailored to gaming and entertainment, thereby facilitating research in AI-powered media delivery and immersive user experiences. The key contributions of this work are summarized as follows.

\begin{figure}[htbp]
\centering
    \includegraphics[width=\linewidth]{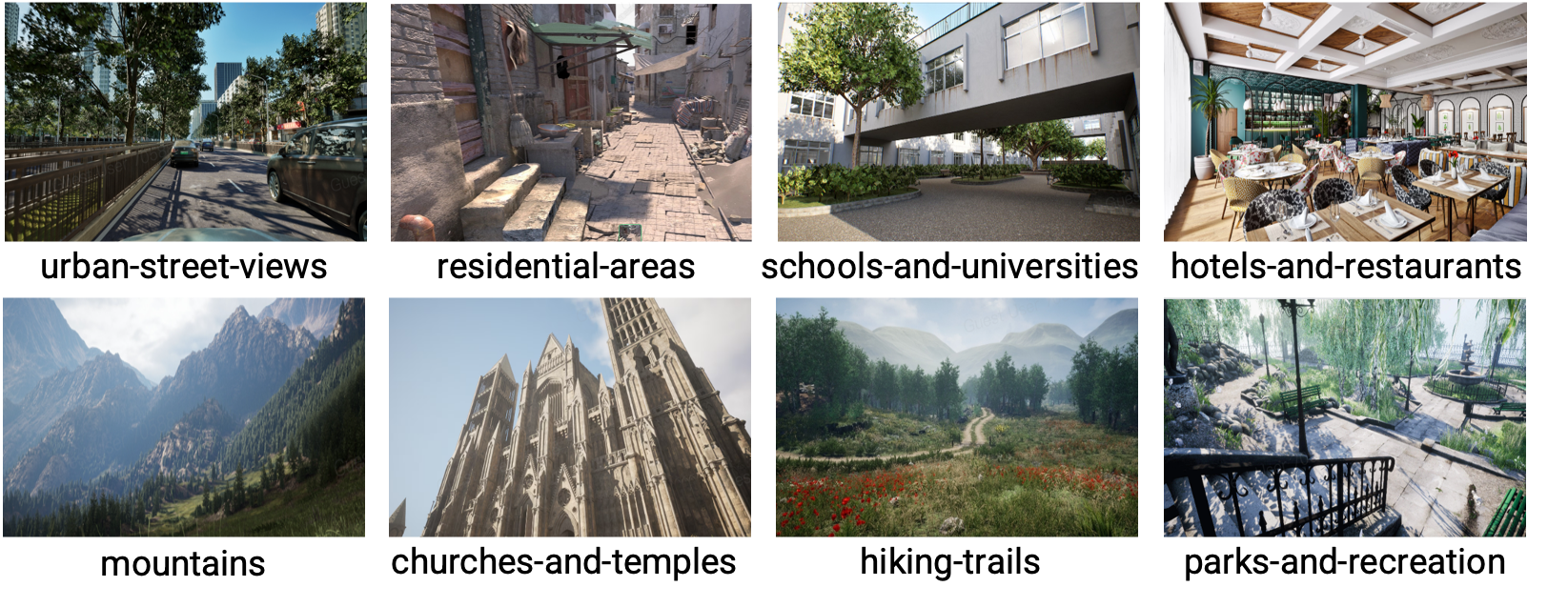}\vspace{-1em}
\caption{Example of 8 scene categories in $\mathtt{M^3VIR}$.}\label{fig:scenecategory}\vspace{-1em}
\end{figure}

\begin{itemize}
\item
{\bf A novel dataset for immersive gaming scenarios.}
Videos in the $\mathtt{M^3VIR}$ dataset have egocentric perspectives and wide fields of view, designed to reflect realistic user scenarios in immersive gaming applications. Compared to previous synthetic datasets \cite{NeRF,Objaverse,D-nerf,GameIR},  $\mathtt{M^3VIR}$ provides 80 scenes across 8 diverse categories (shown in Fig.~\ref{fig:scenecategory}) simulated with Unreal Engine 5 (UE5), with synchronized RGB frames, depth maps, segmentation maps, and associated camera intrinsic and extrinsic parameters. $\mathtt{M^3VIR}$ can serve as ground truth for multiple restoration and content creation tasks.\vspace{.5em}

\item{\bf A multi-resolution subset for media delivery research.}
We introduce $\mathtt{M^3VIR\_MR}$ to support three modern media delivery solutions: Super-Resolution (SR), where LR frames are transmitted and HR frames are reconstructed on the client side; Novel View Synthesis (NVS), where only a subset of views is transmitted and the remaining views are synthesized by the client; and their combination (NVS+SR), where LR frames from partial views are sent and all HR frames are generated locally. This subset comprises 43,200 data samples from 1,440 video sets, with each sample including temporally synchronized RGB frames, segmentation maps, and depth maps at three resolutions, as well as intrinsic and extrinsic camera parameters.\vspace{.5em}

\item{\bf Baseline evaluations for SR, NVS, and NVS+SR}
We benchmark several state-of-the-art (SOTA) algorithms on $\mathtt{M^3VIR\_MR}$. For SR, we evaluate RealESRGAN \cite{Realesrgan}, DAT \cite{DAT}, and ResShift \cite{Resshift} as representative methods based on GANs, transformers, and diffusion models, respectively. For NVS, we test both the NeRF-based NeRFacto \cite{nerfstudio} and 3D Gaussian Splatting (3DGS) \cite{Splatting}. For the combined NVS+SR, we assess the Sequence Matters (SeqMat) framework \cite{SeqMat}, which aims to enhance 3DGS-based NVS with temporal-aware SR to improve temporal consistency. Within the SeqMat pipeline, we further evaluate two representative SR approaches: the transformer-based PSRT \cite{PSRT} as a video super-resolution (VSR) method used by the original SeqMat, and the image-based SwinIR \cite{swinIR}. Our baseline evaluation provides insights into current model performance on real gaming data and identifies key directions for future improvements.

\vspace{.5em}

\item{\bf A multi-style subset for controlled video generation.}
We present $\mathtt{M^3VIR\_MS}$, a subset designed to enable research on controllable video generation. In this subset, videos are rendered in three distinct styles—realistic, cartoon, and metallic—while preserving identical geometry. The style variations  target 10 object categories, ensuring meaningful and localized appearance changes. The corresponding segmentation map, depth map,
and camera intrinsic and extrinsic parameters are also provided. 
Although no existing controllable video generation methods are directly applicable for evaluation on this dataset,  $\mathtt{M^3VIR\_MS}$ serves as a valuable benchmark to support future studies on generating style-consistent and spatial-temporally coherent content under controlled conditions.
\end{itemize}

To the best of our knowledge, $\mathtt{M^3VIR}$ is the first large-scale video dataset with diverse computer-synthesized content that provides ground-truth LR-HR paired and multiview frames at the scene level, along with associated depth maps, segmentation maps, and camera parameters. Furthermore, it is the first dataset to include multi-style, object-level ground-truth with consistent geometry across styles, enabling research on controllable video generation with fine-grained style variations. This comprehensive design makes $\mathtt{M^3VIR}$ as a valuable benchmark for a wide range of vision tasks specifically tailored to gaming content.

\begin{figure*}[htbp]
\centering
    \includegraphics[width=\textwidth]{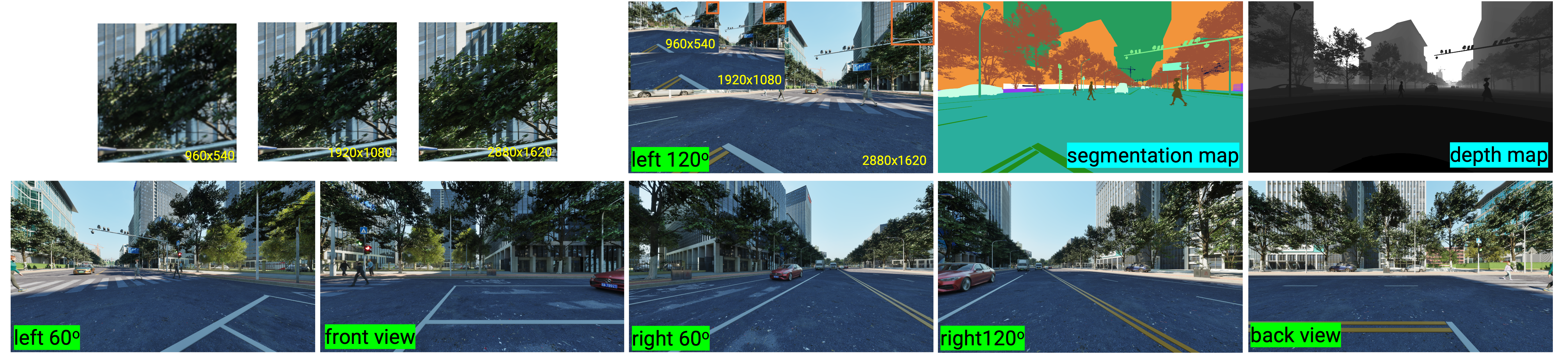}\vspace{-1em}
\caption{One data sample example of $\mathtt{M^3VIR\_MR}$. All 6 views have multi-resolution videos. All RGB frames have associated segmentation maps, depth maps and camera intrinsic and extrinsic parameters (details of only one frame is shown).}\label{fig:mrexample}\vspace{-.5em}
\end{figure*}

\begin{figure*}[htbp]
\centering
    \includegraphics[width=\textwidth]{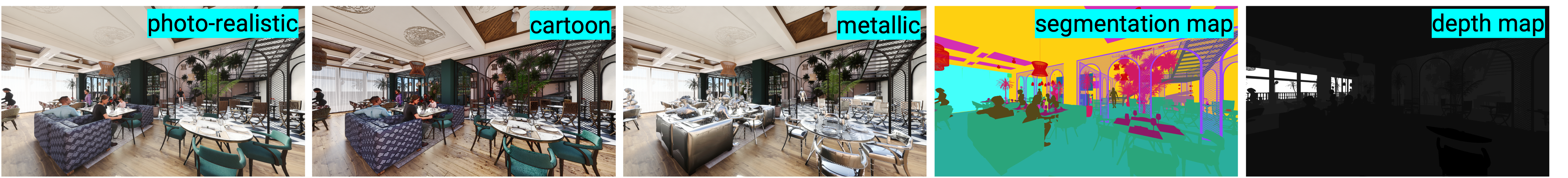}\vspace{-1em}
\caption{A data sample of $\mathtt{M^3VIR\_MS}$ where objects in the scene are changed to different styles.}\label{fig:msexample}\vspace{-1em}
\end{figure*}

\section{Related Works}

\subsection{Super-Resolution: Methods and Datasets}
The pioneering work of SRCNN \cite{SRCNN1, SRCNN2} has inspired extensive research on deep-learning-based single image super-resolution (SISR). To handle the complex real degradations, blind SISR \cite{michaeli2013nonparametric, zhang2018learning} has become the research focus. Methods like \cite{zhang2018learning, gu2019blind, zhang2020deep, maeda2022image} explicitly model and estimate the degradation process, and perform reconstruction based on the estimated degradation model. However, real-world degradation is too complicated to model accurately through simple combinations of multiple degradations. In comparison, implicit methods \cite{ESRGAN, Realesrgan, AdaCode, BSRGAN, Anime4K} automatically learn and adapt to various degradation conditions based on LR training data distribution. Transformer-based and diffusion-based methods have recently outperformed CNN-based and GAN-based methods. Transformer-based SR models, such as SwinIR \cite{swinIR}, HAT \cite{HAT}, PSRT \cite{PSRT}, and Restormer \cite{Restormer}, leverage self-attention to capture long-range dependencies and global context effectively. By employing hierarchical and window-based attention mechanisms, these models balance computational efficiency with superior reconstruction quality. Diffusion-based SR models, such as SR3 \cite{SR3}, ResShift \cite{Resshift}, and StableSR \cite{stableSR} generate HR images via iterative denoising, progressively refining outputs from noise to clean reconstructions. These methods show strong robustness to degradation. 

Although implicit methods have achieved large improvements over real-world images, their performance is highly limited by the training degradations, making them difficult to generalize to out-of-distribution images. Previous methods mitigate this issue by increasing the variety of training degradation types and scales. However, such a training strategy does not work well for gaming content. Real-rendered LR images are clear, sharp, without blur or artifacts, quite different from pseudo LR images generated by applying degradations. As a result, SISR models need to be trained on real LR-HR paired gaming data to learn true degenerative features and improve their performance.

Existing datasets for SISR mainly consist of HR images. Commonly used datasets include DIV2K \cite{DIV2K}, DIV8K \cite{DIV8K}, Flickr2K \cite{Flickr2K}, and Flickr-Faces-HQ (FFHQ) \cite{FFHQ}. Other popular datasets for general vision tasks, such as ImageNet \cite{ImageNet} and COCO \cite{COCO}, are also used for SISR. LR images for these datasets are generated by applying degradations to the HR images. To improve the generalization of models when applied to real-world scenarios, complex degradations have been employed, such as multiple simulated degradations \cite{elad1997restoration} and BSRGAN generated degradations \cite{BSRGAN}. However, the gap between the simulated and real degradations still exists. The problem is especially prominent for gaming images due to their unique characteristics different from natural images. There are some datasets providing real-world ground-truth image pairs, \textit{e.g.}, City100 \cite{City100}, RealSR \cite{RealSR}, and DRealSR \cite{DRealSR}, by using two calibrated devices with varying focal lengths to directly capture LR-HR image pairs. However, due to the expensive process, scale and content diversity is usually highly limited. Also, time synchronization and pixel-level alignment still remain challenging. 

The synthetic GameIR dataset \cite{GameIR} provides rendered (with Unreal Engine UE4) LR-HR paired ground-truth frames from synchronized multiple views, similar to our $\mathtt{M^3VIR}$ dataset. However, It only focuses on self-driven scenes with limited content variation, scene complexity, and object diversity.

\subsection{Novel View Synthesis: Methods and Datasets}

NVS aims to generate novel view images by integrating image data from multiple camera perspectives. Methods based on Neural Radiance Fields (NeRF) \cite{NeRF} have shown great performance over a large variety of scenes.  NeFR++ \cite{NeRF++} builds upon NeRF with improved representations and volume rendering. Mip-NeRF \cite{Mip-NeRF} uses conical frustums rendering to reduce aliasing and enhance applicability to multiscale and high-resolution scenes. Instant-NGP \cite{Instant-NGP} uses hash tables and multi-resolution grids to speed up training and inference. DSNeRF \cite{DSNeRF} leverages depth information for supervision to improve performance. PyNeRF\cite{pynerf} enhances the rendering speed and quality by training models across various spatial grid resolutions.

Recently, 3D Gaussian Splatting (3DGS) \cite{Splatting} 
has gained significant attention for NVS due to its ability to achieve both high rendering quality and real-time performance. Unlike traditional NeRF-based methods that rely on dense neural volumetric rendering and are computationally intensive, 3DGS represents scenes as a set of anisotropic Gaussian primitives with learned properties such as position, color, and opacity. These primitives can be efficiently splatted onto the image plane, enabling rapid view interpolation while preserving fine geometric and appearance details. The approach not only accelerates rendering but also reduces memory requirements, making it highly suitable for interactive applications such as virtual reality, gaming, and real-time scene exploration. 

NVS datasets are generally divided into synthetic and real-world.   Earlier real-world datasets were developed for multi-view stereo tasks, such as Tanks and Temples \cite{tanks-and-temples} and DTU \cite{DTU}, offering limited scene variety. ScanNet \cite{Scannet} contains 3D scans and RGB-D video data, but with motion blur and narrow field-of-view. Later datasets featuring outward-facing and forward-facing scenes have limited diversity in general. For instance, LLFF \cite{llff} provides 24 cellphone-captured forward-facing scenes. Mip-NeRF 360 \cite{Mip-NeRF} provides 9 indoor and outdoor scenes with uniform distance around central subjects. Mill 19 \cite{Mill-19} provides 2 industrial and open-space scenes. 
BlendedMVS \cite{Blendedmvs} offers multi-view images and depth maps but with limited scenes. Recently, large-scale scene-level real-world datasets have emerged. For example, RealEstate10K \cite{RealEstate10K} offers diverse indoor scenes through real estate videos, but with low-resolution and inconsistent quality. Replica \cite{Replica} provides high-quality data including RGB images, depth maps, and semantic annotations, but is limited to indoor environments only. The most recent DL3DV-10K \cite{DL3DV-10K} significantly enriched the real-world scene collection by providing 10,510 videos captured from 65 types of scene locations, with different levels of reflection, transparency, and lighting conditions. 

In comparison, there is a lack of large-scale scene-level synthetic datasets for NVS research over synthetic gaming data. Most existing synthetic datasets are at the object level. For instance, 
Blender \cite{NeRF}, Objaverse \cite{Objaverse}, and D-NeRF \cite{D-nerf}  contain 3D CAD models with varied textures and geometries without real-world noises or non-ideal conditions. The recent GameIR dataset \cite{GameIR} provides synchronously rendered multi-view video frames, but with limited scene variety and complexity. Similar to the super-resolution task, due to the unique characteristics of gaming content, \textit{e.g.}, unnatural object motion with limited motion blur, NVS methods need to be trained and evaluated over diverse scene-level synthetic datasets to assess their effectiveness for gaming data.

\subsection{Video Generation: Methods and Datasets}

Video generation has rapidly advanced with the development of deep generative models. Early methods such as Video Pixel Networks \cite{Kalchbrenner2017} and RNN-based models generate videos frame-by-frame, and often struggle with temporal coherence. GAN-based methods use adversarial training for sharper visuals, including VideoGAN \cite{Vondrick2016} and MoCoGAN \cite{Tulyakov2018}, which disentangle motion and content for controllable video synthesis. Progressive VGAN \cite{Tian2021} improves temporal modeling by leveraging temporal generators. More recently, transformer-based methods such as VideoGPT \cite{Yan2021} and NUWA \cite{Wu2021} have shown strong scalability and quality. Diffusion models, including Video Diffusion Models \cite{Ho2022} and VideoCrafter \cite{Chen2023}, gives SOTA performance for temporally coherent and high-fidelity videos. Text-to-video models like CogVideo \cite{Hong2022}, Make-A-Video \cite{Singer2022}, Nvidia's text-to-video \cite{Cosmos}, and Sora \cite{OpenAI2024} show the potential of large multimodal models to generate videos from text, combining language understanding with generative modeling. 

Despite significant progress in video generation, controlling the generated content using text prompts or image examples remains a major challenge. One key difficulty lies in ensuring semantic alignment between the input condition (e.g., a textual description or reference image) and the generated video, especially when capturing fine-grained details, complex actions, or specific temporal dynamics. Text-to-video models can generate visually plausible videos from prompts, but often struggle with consistency or object continuity across frames. Image-to-video models attempt to preserve appearance and style, but controlling how these features evolve temporally is still limited. Multimodal models tend to overfit to common training patterns, leading to generic or repetitive motions, and they often lack the ability to follow user intent accurately. Improving controllability and user-guided generation—especially over long durations and complex scenes—remains an open and active area of research.

Current video generation datasets lack diversity and detailed control annotations. UCF-101 \cite{Soomro2012}, Kinetics-600 \cite{Kay2017}, and BAIR Robot Pushing \cite{Ebert2017} focus on narrow domains like action recognition or robotic motion. Larger-scale datasets such as WebVid-10M \cite{Bain2021} offer broader coverage but suffer from noisy captions and limited fine-grained control signals, making comprehensive evaluation of controllable generation difficult.

Moreover, evaluating open-ended video generation is challenging due to the absence of ground-truth outputs tied to input prompts. Unlike deterministic video prediction, there is no single correct output, making traditional metrics like PSNR and SSIM inadequate. Learned metrics such as FVD \cite{Unterthiner2019} and CLIPScore \cite{UHessel2021} better reflect semantic similarity but still do not measure fine-grained aspects like temporal consistency or user intent alignment. 

\section{$\mathtt{M^3VIR}$ Dataset}

The proposed $\mathtt{M^3VIR}$ is a large-scale multi-modality, multi-view dataset with egocentric perspectives and wide fields of view, designed to reflect realistic user scenarios in immersive gaming applications.
$\mathtt{M^3VIR}$ includes 80 scenes across 8 categories, 10 scenes per category:  \textit{urban-street-views}, \textit{residential-areas}, \textit{school-universities},  \textit{hotels-and-restaurants}, \textit{mountains}, \textit{churches-and-temples}, \textit{hiking-trails}, \textit{parks-and-recreation-areas} (examples shown in Fig.~\ref{fig:scenecategory}). A variety of videos are simulated with consistent scene content using the Unreal Engine 5 (UE5) to serve as ground truth for multiple vision tasks, such as restoration and controlled content generation.

\subsection{The $\mathtt{M^3VIR\_MR}$ Subset}
Within $\mathtt{M^3VIR}$ we provide $\mathtt{M^3VIR\_MR}$, a multi-resolution subset, to support three representative media delivery solutions. The first is SR, where the server renders and transmits only LR images, and HR images are reconstructed on the client side. The second is NVS, where the server sends images from only a subset of views, and the client synthesizes the remaining views. The third is a combined approach, NVS+SR, in which the server transmits LR images from a subset of views, and the client device generates  HR images of all views. This subset is designed to simulate practical delivery scenarios and support the development of efficient, AI-driven cloud gaming solutions.

Specifically, for each of the 80 scenes, 3 sets of multi-modal multi-view data packages are collected: dynamic scene with static camera; static scene with moving camera; dynamic scene with moving camera. Each set of data package comprises 6 sets of temporally synchronized RGB videos from co-located cameras with 6 views. Each of the 6 sets further has videos at 3 different resolutions: $960\!\times\!540$, $1920\!\times\!1080$, $2880\!\times\!1620$. In addition to RGB images, the corresponding pixel-level synchronized semantic segmentation map and depth map are also provided. Each video is 2-sec long at 15fps. In total, $\mathtt{M^3VIR\_MR}$ has 43200 data samples from 1440 sets of videos, each data sample consisting of matching RGB images, segmentation maps and depth maps at 3 different resolutions. The corresponding intrinsic and 6-DoF extrinsic camera parameters for each frame are also provided. Fig.\ref{fig:mrexample} shows an example of one data sample in $\mathtt{M^3VIR\_MR}$.

\subsection{The $\mathtt{M^3VIR\_MS}$ Subset}
Within $\mathtt{M^3VIR}$, we provide a multi-style subset, $\mathtt{M^3VIR\_{MS}}$, to support research on controlled video generation. Specifically, from the 43200 data samples in $\mathtt{M^3VIR\_MR}$, for the video with $1920\!\times\!1080$ resolution, we render a cartoon-style video and a metallic-style video with the same geometry. Overall, $\mathtt{M^3VIR\_MS}$ contains 43200 data samples, each comprising 3 videos having matching geometry at the frame level and with 3 styles (photo-realistic, cartoon, and metallic), as well as the corresponding segmentation map, depth map, and camera intrinsic and extrinsic parameters. The style change focuses on 10 object categories: people, animals, cars, trees, buildings, mountains, water-surface, tables, coach-chairs, lights-lamps. Fig.~\ref{fig:msexample} gives an example.  

\section{Challenge Tracks}

The ground-truth LR-HR paired frames in $\mathtt{M^3VIR\_MR}$ are used to facilitate restoration research in track $1\sim 3$. 

\subsection{\bf{Track 1: SR}}

To support the media delivery solution of transferring LR frames and restoring HR frames by client. The segmentation and depth map can be leveraged to enhance performance.  $\mathtt{M^3VIR\_MR}$ supports $2\times$ and $3\times$ SR: from $960\!\times\!540$ to $1920\!\times\!1080$ and to $2880\!\times\!1620$. 

\subsection{\bf{Track 2: NVS}}

To support the solution of transferring part of multi-view frames and generating the remaining frames by client. The task is to synthesize intermediate RGB frames from a sparse set of reference RGB frames in multi-view videos. Only 80 sets of videos (6 videos per set for 6 views) for static scenes with $1920\!\times\!1080$ resolution are used for this track.

\subsection{\bf{Track 3: NVS+SR}}

A combination of track 1 \& 2 to support the solution of transferring part of multi-view LR frames and generating all HR frames by client. Track 3 uses the same 80 sets videos as track 2, but with all resolutions to support $2\times$ and $3\times$ SR.

\subsection{Track 4: Object Style Transfer in Video} 

$\mathtt{M^3VIR\_MS}$ provides ground-truth to study controlled video generation in track 4. The target is to edit specific objects in a photo-realistic video by spatial-temporal consistently changing the style of objects (to cartoon style or to metallic style). We focus on 10 object categories: people, animals, cars, trees, buildings, mountains, water-surface, tables, coach-chairs, lights-lamps. $\mathtt{M^3VIR\_MS}$ enables training and evaluating content editing methods with ground-truth paired data. To reduce the difficulty of this challenging task and accommodate different possible solutions, only 14400 data samples corresponding to the static scenes are used for evaluation. 

\section{Baseline Evaluation}

We evaluated SOTA algorithms over the $\mathtt{M^3VIR}$ dataset for track 1, 2, and 3 in two ways: evaluating pretrained models, evaluating finetuned models using the $\mathtt{M^3VIR}$ training set. 

\subsection{Evaluated SR Methods for Track 1}

We tested 3 representative SR methods: Real-ESRGAN \cite{Realesrgan}, DAT \cite{DAT} and ResShift \cite{Resshift}. Real-ESRGAN is a widely adopted GAN-based approach designed to handle diverse real-world image degradations, giving sharp and perceptually pleasing reconstructions. DAT is a transformer-based method that alternates spatial and channel self-attention to capture both global context and fine details, by using the adaptive interaction module (AIM) for effective feature exchange between branches and the spatial-gate feed-forward network (SGFN) to enhance spatial awareness. DAT often gives SOTA SR performance with sharp, high-fidelity reconstructions. ResShift is a diffusion-based SR framework that accelerates the denoising process by directly shifting LR-HR residuals with a customized noise schedule. Each of these methods demonstrates unique strengths across different application scenarios and represents the current SOTA in its respective category. By evaluating both their pretrained and fine-tuned models, we gain a good understanding of how modern SR techniques perform on rendered gaming content.

\subsubsection{Implementation Details.} For each scene category in $\mathtt{M^3VIR\_MR}$, 8 scenes were randomly selected to form the training set, the remaining 2 for testing. This ensured distinct scene in training and test data. For SR evaluation, all sets of multi-modal multi-view data packages were used: dynamic scene with static camera; static scene with moving camera; dynamic scene with moving camera. We tested two SR settings: $\times 2$ SR from $960\!\times\!540$ to $1920\!
\times\!1080$ resolution, and $\times 3$ SR from $960\!\times\!540$  to $2880\!\times\!1620$ resolution. 

For Real-ESRGAN and DAT, we directly used their published source code and pretrained models without any finetuning. In contrast, ResShift only provides public checkpoints for the default $\times 4$ SR setting. To establish fair baselines for $\times 2$ and $\times 3$ SR, we (i) trained a new model using the official $\times 2$ configuration from ResShift, and (ii) created an in-house $\times 3$ variant by scaling the parameters and finetuning it on the same ImageNet training split. All experiments strictly followed the methodologies and hyperparameters specified in the ResShift papers. To ensure fairness and reproducibility, our training used eight V100 GPUs, and all test evaluations are conducted on a single V100 GPU.

\subsection{Evaluated NVS Methods for Track 2}

We evaluated NVS methods based on both NeRF (NeRFacto\cite{nerfstudio}) and 3DGS \cite{Splatting}. NeRFacto generates ray bundles by optimizing camera views and employs segmental samplers for ray sampling. It also combines scene contraction with appearance embedding, providing SOTA efficiency of NeRF models in complex scenes. 3DGS represents scenes as anisotropic Gaussian primitives that are efficiently splatted onto the image plane, providing an excellent balance between rendering speed and visual fidelity. 
Each method has strengths in different applications, and evaluating them on $\mathtt{M^3VIR}$ allows us to assess the performance of leading NVS techniques in gaming scenarios.

\subsubsection{Implementation Details.}
The overall train/test split for NVS was the same as that for SR above, \textit{i.e.}, the same 2 scenes in each scene category were used for testing. From the data package for each test scene, only the video corresponding to the static scene with moving camera with $1920\!\times\!1080$ resolution was used for NVS evaluation. This reasonably simplified the NVS task to focus on testing backbone general NVS methods, since NVS of dynamic scene remains a challenging open problem in the field. Specifically, for each test video, we randomly sampled $n\%$ frames for training the NeRF or 3DGS model, and synthesized the remaining $(100-n)\%$ frames for evaluation.  

We used the NeRFStudio implementation \cite{nerfstudio} of NeRFacto, and the original published source code for 3DGS \cite{Splatting}. The default provided hyperparameters were used.  We tested performance with different sampling rates $n$: 80\%, 70\%, and 60\%. For all models, the training and testing were done on a single V100 graphics card.

\subsection{Evaluated NVS+SR Methods for Track 3}

For Track 3 we adopted the pipeline of Sequence Matters (SeqMat \cite{SeqMat}), which is the first framework that fuses off-the-shelf Video Super Resolution (VSR) networks with 3DGS to improve both texture fidelity and cross-view consistency, outperforming single-image SR or naive NVS approaches. Specifically, SeqMat first applies VSR and then feeds the enhanced sequence, together with the original camera parameters, to a 3DGS optimiser.

The original SeqMat uses PSRT \cite{PSRT}, a transformer-based VSR model featuring a patch alignment mechanism that coarsely aligns image patches instead of relying on expensive pixel-wise warping. This design allows it to leverage multi-frame cues efficiently while remaining robust to large motions. To evaluate the impact of temporal modeling, we also tested SwinIR \cite{swinIR}, a strong image-based SR method, where each frame was super-resolved independently before being processed by 3DGS. Importantly, our $\mathtt{M^3VIR}$ dataset provides sub-pixel accurate ground-truth camera poses, eliminating discrepancies caused by pose estimation errors. Comparing these SR variants within the same SeqMat+3DGS pipeline enables a clear, quantitative evaluation of how video-aware versus image-centric SR strategies impact texture realism and cross-view consistency in game-style 3D reconstructions.

\subsubsection{Implementation Details.}

Same as Track 1, we evaluated both $\times 2$ and $\times 3$ SR configurations: from $960\!\times\!540$ to $1920\!\times\!1080$ and from $960\!\times\!540$ to $2880\!\times\!1620$. The ground-truth camera poses were directly fed into the SeqMat pipeline. For PSRT, the public $PSRT$\_$Vimeo$ checkpoint was used, and for SwinIR, the classical $DF2K$ checkpoint was used. After SR, the SeqMat pipeline used Adaptive-Length Sequencing (ALS) to cluster the upsampled frames into subsequences of at most eight images according to similarity computed based on camera poses and visual features. Then the clustered subsequences, together with the camera poses are fed into the 3DGS optimiser. For all methods, 3DGS ran 30k steps on a single NVIDIA V100. 

Specifically, SeqMat leverages both HR and LR loss terms during 3DGS training. To enhance texture sharpness and local contrast, the rendered image is compared against the upscaled frame. At the same time, the rendered image is downsampled and compared with the original LR frame to maintain fidelity to the source. By jointly optimizing these complementary losses, SeqMat produces upscaled frames with sharp details while preserving global consistency relative to the LR input video.

\section{Evaluation Results}
\label{sec:experiments}
We evaluated visual quality using objective metrics including PSNR, SSIM, LPIPS, FID, and DISTS. Traditional metrics PSNR and SSIM measure pixel-level restoration accuracy, assessing how closely reconstructed images match the ground truth. In contrast, perceptual metrics—LPIPS, FID, and DISTS—focus on visual quality as perceived by humans. LPIPS evaluates local perceptual differences, capturing sharpness and fine details. FID measures the distributional distance between generated and real images, reflecting overall perceptual realism. DISTS combines structure and texture similarity using deep CNN features, offering better alignment with human visual perception. Together, these metrics provide complementary insights into both fidelity and perceptual quality for restoration and generation tasks.

\subsection{SR Results for Track 1}

Tab. \ref{tab:realesrgan_metrics} summarizes the single-frame SR results on $\mathtt{M^3VIR}$. Across all metrics, the transformer-based DAT consistently outperforms the GAN-based Real-ESRGAN and the diffusion-based ResShift. For the $\times 2$ SR task, DAT demonstrates a clear advantage in PSNR and SSIM, reflecting superior reconstruction fidelity, while also achieving slightly better perceptual quality scores. In the more challenging $\times 3$ SR setting, DAT maintains its lead in PSNR and SSIM, though its perceptual quality becomes comparable to ResShift across LPIPS, FID, and DISTS. Real-ESRGAN shows competitive LPIPS and DISTS values but performs significantly worse in FID, suggesting inconsistencies in how FID correlates with other perceptual metrics over synthetic content. Overall, transformer-based and diffusion-based methods achieve better results on synthetic gaming data compared to the GAN-based approach, which is aligned with observations from SR research on natural images.

\begin{table}[htbp]
\caption{SR baseline performance for Track 1 }
\centering
\scalebox{0.8}{%
\begin{tabular}{lcccccc}
    \toprule
    Method & Scale & PSNR $\uparrow$ & SSIM $\uparrow$ & LPIPS $\downarrow$ & FID $\downarrow$ & DISTS $\downarrow$\\
    \midrule
    Real-ESRGAN & x2 & 24.75 & 0.78 & 0.11 & 43.3038 & 0.2036\\
    Real-ESRGAN & x3 & 23.90 & 0.73 & \bf{0.15} & 51.8325 & 0.2041\\
    DAT & x2 & \bf{28.51} & \bf{0.86} & \bf{0.09} & \bf{21.4278} & \bf{0.1943}\\
    DAT & x3 & \bf{26.85} & \bf{0.81} & 0.18 & 29.1548 & \bf{0.1944}\\
    ResShift & x2 & 26.60 & 0.78 & \bf{0.09} & 23.0597 & 0.1987\\
    ResShift & x3 & 25.22 & 0.73 & 0.16 & \bf{24.6438} & 0.2005\\
    \bottomrule
\end{tabular}%
}
\label{tab:realesrgan_metrics}
\end{table}

\begin{figure*}[htbp]
\centering
    \includegraphics[width=\textwidth]{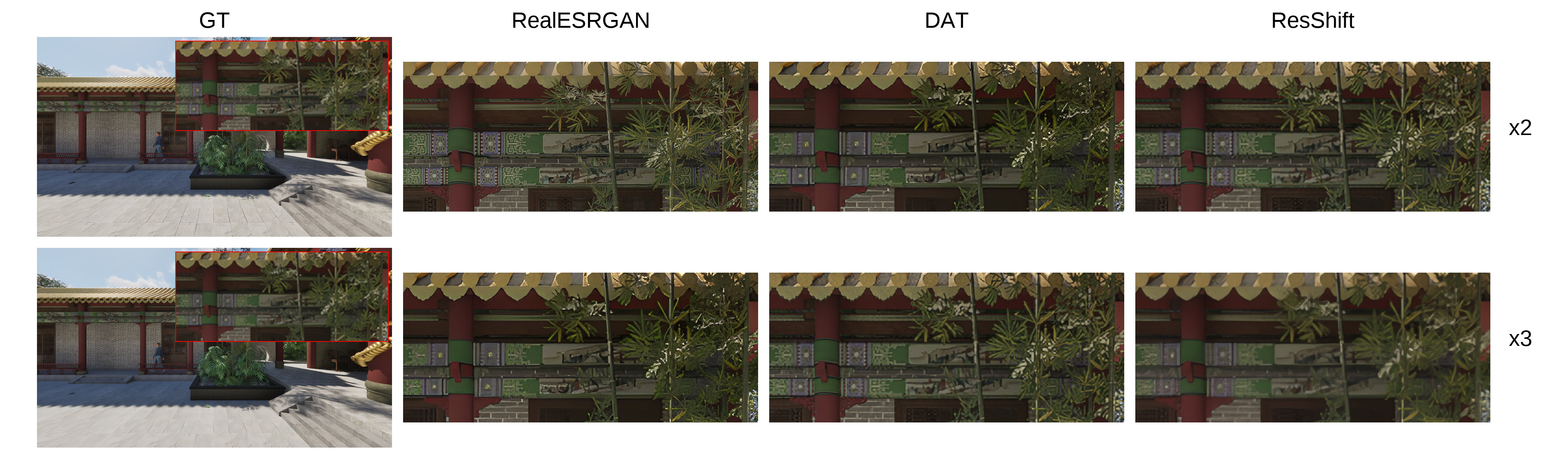}
\caption{Qualitative comparison examples of Real-ESRGAN, DAT, and ResShift for Track 1}\label{fig:track1_examples}
\end{figure*}

Fig.~\ref{fig:track1_examples} shows qualitative comparison examples for Track 1. Overall, DAT reconstructs sharp and natural details that closely match the ground truth, as seen in fine structures like bamboo leaves and lattice patterns. Real-ESRGAN often introduces ringing artifacts along edges and loses finer details, while ResShift tends to produce overly smooth regions with softened textures, reducing the realism of intricate features.

\subsection{NVS Results for Track 2}

We evaluated NeRFacto and 3DGS on $\mathtt{M^3VIR_MR}$ using three train/test splits: $80\%/20\%$, $70\%/30\%$, and $60\%/40\%$ of randomly sampled frames for training and testing, respectively. As shown in Table~\ref{tab:recon}, 3DGS consistently outperforms NeRFacto across all metrics. Notably, even with only $60\%$ of frames used for training, 3DGS achieves reconstruction quality comparable to the $80\%$ training setting, confirming its data efficiency and robustness to sparse training views. 

An additional observation from Table~\ref{tab:recon} is the inconsistency among perceptual quality metrics. Similar to Track 1, where FID showed weak correlation with other perceptual measures, DISTS also exhibits inconsistent correlation with other perceptual metrics in Track 2. This highlights the need for further research on developing more reliable evaluation metrics for perceptual quality.

\begin{figure*}[htbp]
\centering
    \includegraphics[width=\textwidth]{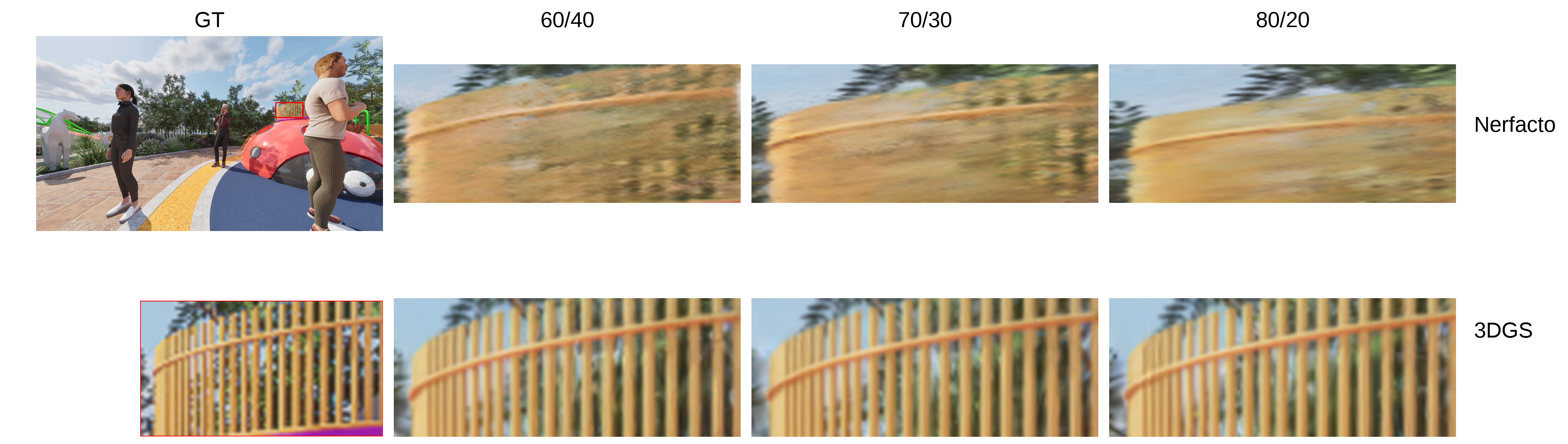}
\caption{Qualitative comparison examples of NeRFacto and 3DGS on different train/test splits for Track 2. }\label{fig:track2_examples}
\end{figure*}

Fig.~\ref{fig:track2_examples} presents qualitative comparisons for Track 2. Across all split settings, 3DGS consistently preserves the sharpness and straightness of fine structures, such as fence bars. In contrast, NeRFacto produces blurry and warped reconstructions, even when trained with a higher proportion of views, demonstrating its limitations in maintaining geometric fidelity.

\begin{table}[htbp]
\caption{NVS baseline performance for Track 2}
\centering
\scalebox{0.8}{%
\begin{tabular}{lccccccc}
    \toprule
    Method & Train (\%) & Test (\%) & PSNR $\uparrow$ & SSIM $\uparrow$ & LPIPS $\downarrow$ & FID $\downarrow$ & DISTS $\downarrow$\\
    \midrule
    NeRFacto & 80 & 20 & 25.89 & 0.76 & 0.20 & 39.9621 & 0.323\\
    NeRFacto & 70 & 30 & 25.32 & 0.75 & 0.21 & 37.4045 & 0.322\\
    NeRFacto & 60 & 40 & 25.85 & 0.76 & 0.20 & 36.1440 & 0.315\\
    3DGS     & 80 & 20 & \bf{33.60} & \bf{0.93} & \bf{0.12} & \bf{17.1503} & \bf{0.2207}\\
    3DGS     & 70 & 30 & \bf{33.51} & \bf{0.90} & \bf{0.13} & \bf{16.4111} & \bf{0.2217}\\
    3DGS     & 60 & 40 & \bf{33.47} & \bf{0.94} & \bf{0.12} & \bf{15.3379} & \bf{0.2238}\\
    \bottomrule
\end{tabular}%
}
\label{tab:recon}
\end{table}

\subsection{NVS + SR Results for Track 3}

We evaluated two variants within the SeqMat framework: PSRT-SeqMat and SwinIR-SeqMat, where PSRT and SwinIR serve as video-based and image-based SR methods, respectively, prior to 3DGS fitting in the SeqMat pipeline. Table~\ref{tab:nvssr} summarizes the performance comparison. Unlike the previous tracks, the evaluated methods exhibit notable inconsistencies across different evaluation metrics. For example, in terms of fidelity, the temporal-aware PSRT achieves higher SSIM but lower PSNR in the $\times 2$ setting, whereas it shows the opposite trend—higher PSNR but lower SSIM—in the $\times 3$ setting. Regarding perceptual quality, PSRT outperforms SwinIR-SeqMat across all perceptual metrics (LPIPS, FID, and DISTS) for the $\times 2$ setting but underperforms in the $\times 3$ setting.

\begin{figure*}[htbp]
\centering
    \includegraphics[width=0.8\linewidth]{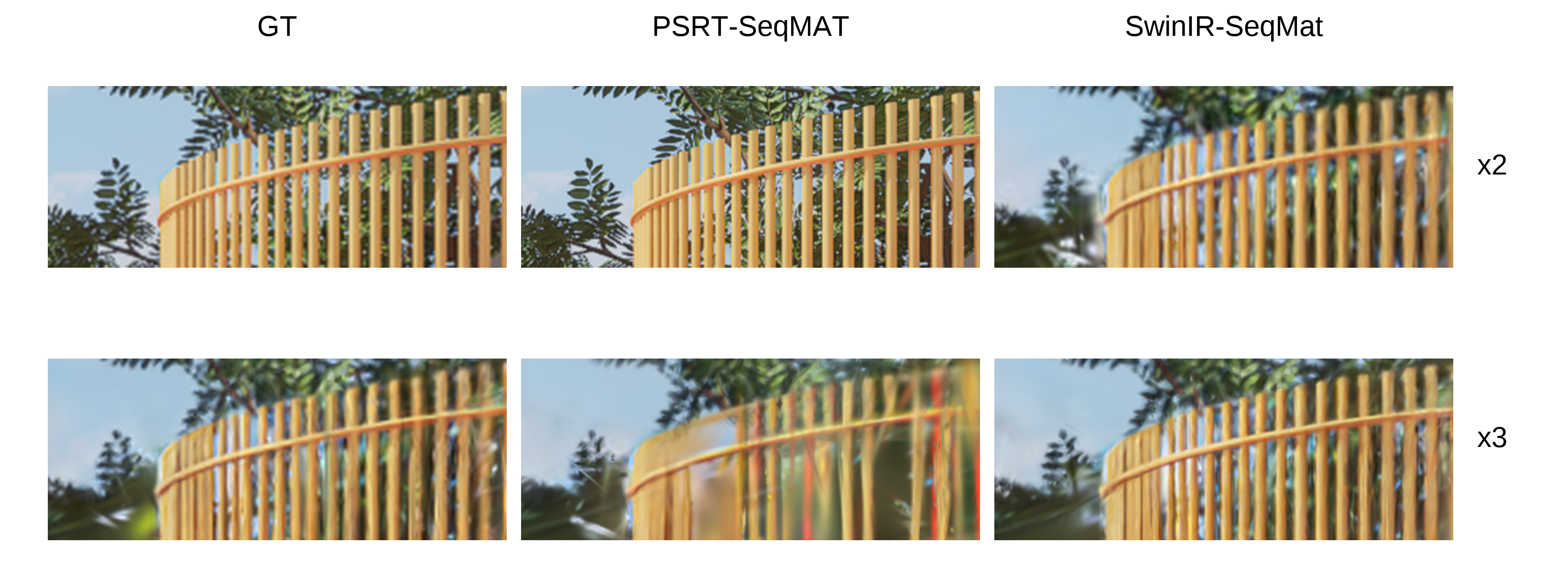}\vspace{-1em}
\caption{Qualitative comparison examples of PSRT-SeqMat and SwinIR-SeqMat for Track 3}\label{fig:track3_examples}
\end{figure*}

Figure~\ref{fig:track3_examples} provides a visual side-by-side comparison that further illustrates these inconsistencies. As shown, SwinIR-SeqMat generally produces sharper reconstructions than PSRT-SeqMat but introduces slight ringing artifacts in the $\times 2$ setting. In contrast, PSRT-SeqMat suffers from severe blurring in the $\times 3$ setting, where structures such as bars appear blended and dramatically distorted. These results suggest that rendered gaming videos exhibit temporal characteristics—such as dynamic lighting and motion variations—that differ significantly from those in natural videos. Such temporal discrepancies, combined with spatial color and texture differences, introduce domain shifts that severely degrade the performance of PSRT models pretrained on natural video datasets.

\begin{table}[htbp]
\caption{NVS + SR baseline performance for Track 3.}
\centering
\scalebox{0.8}{%
\begin{tabular}{lcccccc}
    \toprule
    Method & Scale & PSNR $\uparrow$ & SSIM $\uparrow$ & LPIPS $\downarrow$ & FID $\downarrow$ & DISTS $\downarrow$\\
    \midrule
    PSRT-SeqMat & x2 & 23.89 & \bf{0.814} & \bf{0.258} & \bf{83.51} & \bf{0.2371}\\
    PSRT-SeqMat & x3 & \bf{23.49} & 0.744 & 0.381 & 132.9696 & 0.2484\\
    SwinIR-SeqMat & x2 & \bf{24.12} & 0.812 & 0.267 & 83.5931 & 0.2384\\
    SwinIR-SeqMat & x3 & 22.95 & \bf{0.75} & \bf{0.325} & \bf{102.0554} & \bf{0.2423}\\
    \bottomrule
\end{tabular}%
}
\label{tab:nvssr}
\end{table}

\subsection{More discussions about Track 4}

We do not provide baseline evaluations for Track 4, as there are currently no existing methods specifically designed for object-level style transfer in controllable video generation. Unlike traditional style transfer techniques that operate at the image or frame level, our $\mathtt{M^3VIR\_MS}$ dataset introduces the challenge of maintaining consistent geometry while applying localized style changes to specific object categories across frames and views. By releasing $\mathtt{M^3VIR\_MS}$ to public, we aim to encourage the research community to explore this underdeveloped area, assisting the development of new methods capable of achieving fine-grained, style-consistent, and temporally coherent video generation.

\section{Conclusion}
We introduced $\mathtt{M^3VIR}$, a large-scale, multi-modality, multi-view dataset specifically designed to facilitate research for gaming and entertainment restoration and content creation. The dataset includes a multi-resolution subset, $\mathtt{M^3VIR\_MR}$, supporting modern media delivery solutions such as SR, NVS, and NVS+SR, as well as a multi-style subset, $\mathtt{M^3VIR\_MS}$, enabling research on controllable video generation with object-level style variations.

Through comprehensive baseline evaluations of SOTA SR, NVS, and NVS+SR methods, we revealed the challenges posed by synthetic gaming data and demonstrated the strengths and limitations of current approaches. Through evaluation we also observed inconsistencies in existing quality metrics, highlighting the need for better evaluation methods.

To the best of our knowledge, $\mathtt{M^3VIR}$ is the first dataset providing diverse content, accurate ground truth, and multi-style object-level annotations, making it a valuable benchmark for advancing research in AI-powered restoration, compression, and controllable content generation.


\bibliographystyle{ACM-Reference-Format}
\bibliography{main}


\end{document}